# A Novel Platform for Internet-based Mobile Robot Systems

P. M. Duong, T. T. Hoang, N. T. T. Van, D. A. Viet and T. Q. Vinh
Department of Electronics and Computer Engineering
University of Engineering and Technology
Vietnam National University, Hanoi

*Abstract*—**In this paper, we introduce a software and hardware structure for on-line mobile robotic systems. The hardware mainly consists of a Multi-Sensor Smart Robot connected to the Internet through 3G mobile network. The system employs a client-server software architecture in which the exchanged data between the client and the server is transmitted through different transport protocols. Autonomous mechanisms such as obstacle avoidance and safe-point achievement are implemented to ensure the robot safety. This architecture is put into operation on the real Internet and the preliminary result is promising. By adopting this structure, it will be very easy to construct an experimental platform for the research on diverse teleoperation topics such as remote control algorithms, interface designs, network protocols and applications etc.**

## I. INTRODUCTION

Seventeen years after the first system appeared in 1994 [3], Internet-based telerobot has made a great contribution to the modern life allowing us to remotely visit museums, tend gardens, navigate undersea, float in blimps, or handle protein crystals[1]–[5]. Whereas early researches tried to answer the question of how to control a robot through the Internet [3][6][7], recent studies have focused on how to control it in real time and deal with the inevitable Internet transmission delay, delay jitter and non-guaranteed bandwidth, etc [8]-[10]. To be effective, a research project may focus on only one single specific aspect and a common experimental platform is usually desirable for the purpose of implemental verification; however, significant work will be needed to build an experimental platform from scratch. Several Internet-based robot platforms therefore have been proposed with their strengths and limitations [9][11][12].

In [11], a web-based telerobot framework is developed in which communications between users and the robot are centered around a web server. The system consists of four modules: a commercial Pioneer mobile robot, a visual feedback display, a global environment map and a web interface, communicated over TCP protocol. By using the web interface, users are able to control the robot over the Internet to explore a laboratory or to push a ball into a goal. The use of TCP which was originally designed for the reliable transmission of static data such as e-mails and files over low-bandwidth, high-error-rate networks as the communication protocol, however, may limit the system from future developments in which the real-time attribute is highly demanded. In addition, the lack of autonomous mechanisms may influence the robot safety and downgrade the system performance in case of network congestion or interruption.

A more flexible and extensible approach is to use client-server architecture as described in [12]. This modular structure allows users to quickly construct further developments of Internet mobile robot. The system uses UDP as the transport protocol and includes essential modules for an Internet robot system such as a mobile robot, a visual feedback display, a virtual environment display and a user-friendly graphic user interface.

On a similar note, Dawei *et al* proposed a quite complete online robot platform in which an Omni-directional mobile robot with a five DOFs arm is controlled over the Internet by using a virtual represent [9]. During the control session, virtual environment of the remote site is continuously updated at the local site and the next position of the robot is predicted and pointed out in the virtual environment. This combined with a visual feedback display enables user to effortlessly navigate the mobile robot in an unknown environment. In addition, the robot safety is strictly ensured by build-in autonomous mechanisms. The use of sonar sensors with a measuring range of 4cm to 400cm for building the virtual environment, however, may limit the applicability of the system to indoor applications only.

In this paper, we propose a novel platform for Internet-based mobile robot systems with improvements in hardware configuration and software development. The system is in Client-Server mode, which contains users, as the command input and the Multi-Sensor Smart Robot (MSSR), as the controlled plant. The MSSR has accurate motion control with PID algorithm and contains many types of sensors to support diverse purposes of development. The MSSR is connected to the Internet via 3G mobile network. Autonomous mechanisms including obstacle avoidance and safe-point achievement are implemented in the robot. The software has two main modules: a client controller at the user site and a server module at the robot site. A multi-protocol model is applied to deliver the data exchanged between the client and server. The platform is implemented on the real Internet and the experimental result is promising. By adopting this, it will be very easy to construct an experimental system for the research on diverse teleoperation topics such as remote control algorithms, interface designs, network protocols and applications etc.

The paper is organized as follows. Details of the hardware configuration is described in Section II. The software development is described in Section III. Section IV introduces experiments in the real Internet environment. The paper concludes with an evaluation of the system, with respect to its strengths and weaknesses, and with suggestions of possible future developments.

## II. HARDWARE DESIGN

In order to be a platform for different developments, the system hardware needs to be designed to support not only a specific task but also a wide range of applications including both indoor and outdoor environments. In our system, the hardware design is split into three perspectives: the communication configuration, the sensor and actuator, and the user interaction; each is developed with the feasibility, the flexibility and the extendibility in mind. Fig.1 shows an overview of the system.

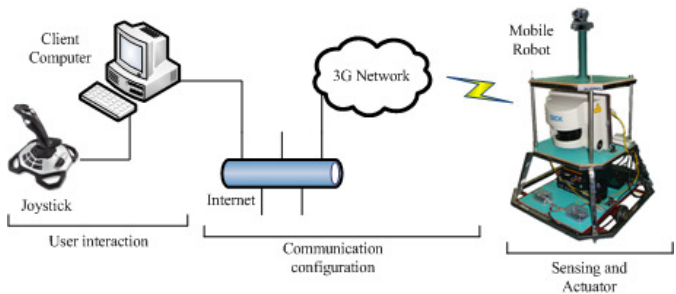

Figure 1. System hardware configuration

### A. Communication configuration

To the best of our knowledge, most current Internet robot systems use a common configuration for the network connection in which the robot and components are connected to the Internet through a central wireless access point (fig.2) [9][11][12]. This configuration is easy to set up but it restricts the operational area of robot and components to a radius of several hundred meters due to the transmit power limitation of the wireless access point. This range is acceptable for indoor environment but is insufficient for outdoor applications such as traffic control and disaster rescue.

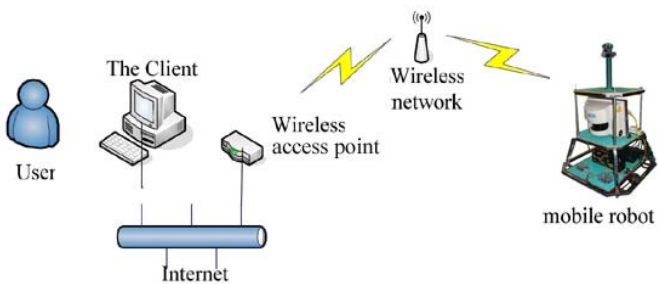

Figure 2. Network configuration of current Internet-based robot systems

In our system, instead of using a wireless access point, the 3G mobile network is utilized as the communicating bridge between the robot and the Internet. A 3G USB device with an internal mobile SIM card is used. The USB is plugged into the computer inside the robot and is registered to a mobile phone service provider allowing it to have access to the Internet (fig.1). This simple configuration enables the robot to connect to the Internet without any restrictions in physical distance as far as the 3G mobile signal is presented which is almost everywhere in the country due to the fact that the 3G signal already covers it all.

### B. Sensors and Actuators

The sensors and actuators are included in a Multi-Sensor Smart Robot (MSSR) developed by our laboratory. The scheme in fig.3 describes sensors, actuators and communication channels in the MSSR. It contains basic components for motion control, sensing and navigation. These components are drive motors for moving control, sonar ranging sensors for obstacle avoidance, compass and GPS sensors for heading and global positioning, and laser range finder (LMS) and visual sensor (camera) for mapping and navigating.

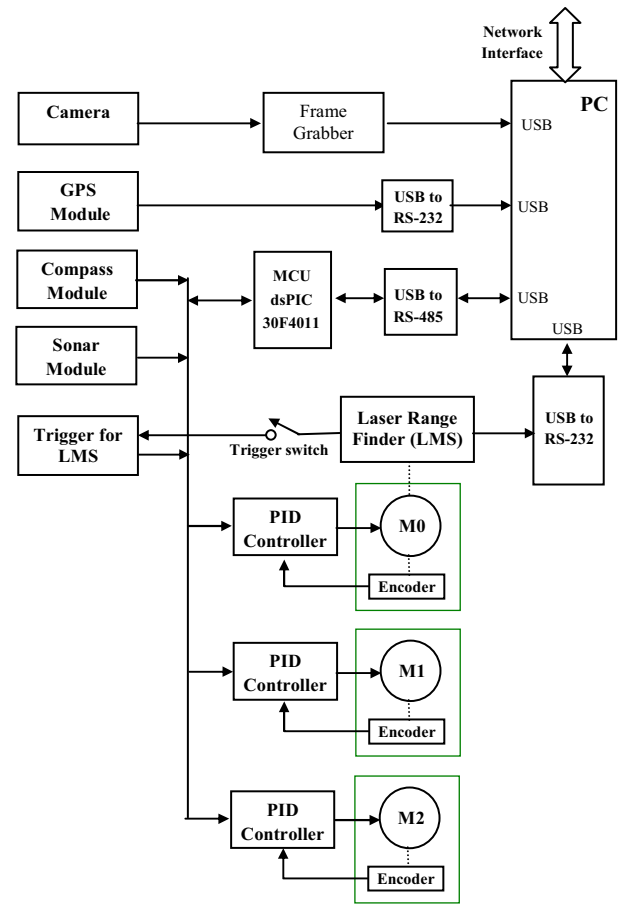

Figure 3. Sensors, actuators and communications in the MSSR

The drive system uses high-speed, high-torque, reversible-DC motors. Each motor is attached a quadrature optical shaft encoder that provides 500 ticks per revolution for precise positioning and speed sensing. The motor control is implemented in a microprocessor-based electronic circuit with an embedded firmware which permits to control the motor by a PID algorithm.

The positioning and heading modules contain a CMPS03 compass sensor and a HOLUX GPS UB-93 module [14][15]. The compass sensor has the good heading resolution of 0.1°. The GPS with lower accuracy is used for positioning in outdoor navigation. Due to the networking is available in our system, an Assisted GPS (A-GPS) can be also used in order to locate and utilize satellite information from the network in poor signal conditions.

The MSSR provides eight SFR-05 ultrasonic sensors split into four arrays, two on each, arranged at four sides of the robot. The measuring range is from 0.04m to 4m.

On the front side of the robot, a 3D-image capturing system is built based on a SICK-LMS 221 2D laser range finder [13][22]. The system has the horizontal view angle of 100° (angle resolutions are 0.25°, 0.5° and 1°) and the vertical (pitching) view angle of 25°. The data produced by ultrasonic sensors and laser range finder which covers a range from 0.04m to 80m is used to build global and local maps of the robot's operational environment.

The visual system is detachable and mounted on the head of the MSSR. It mainly consists of a Sony EVI-D100 pan-tilt-zoom (PTZ) color camera and an EasyCap adapter, which is to capture the video. The rotation ranges of the pan-tilt camera are from -100 to +100 degrees in horizon, from -25 to +25 degrees in vertical and are available to give the user a clear view of the environment in front of the robot.

The communication data between devices and the computer in the robot is transferred via several channels: low-rate channels with standards of RS-485 and RS-232 and high-rate channels with USB ports. Devices using the RS-485 are managed by an on-board 60MHz Microchip dsPIC30F4011-based microcontroller with independent controller boards for a versatile operating environment. A RS-485 bus is established to maintain the communication between controllers and reserve the expansibility to support various accessories. Devices using the RS-232 are directly connected the USB-to-COM modules. Commands of control and acquisition with short messages are realized in low-rate channels. On the other side, images from camera are captured by a frame grabber and directed to a high-rate USB port. The communication between the remote-site (robot) and client-site (user), as described previously, is realized by computer network.

*C. User interaction devices*

The interaction devices at the user site simply consist of a personal computer and a joystick. The computer is an ASUS notebook computer with 1.5GHz M-Centrino processor, 500Mb RAM and Windows XP operating system. The computer, with installed control software, allows users to retrieve feedback information of the remote site and navigate the robot to explore an unknown environment. To support users with a more convenient way of control, a joystick is added. The joystick is the 3D Logitech Extreme series with 10 bit resolution in horizontal and vertical axes and 12 functioning buttons. In the system, the joystick interprets users' inputs to a sequence of control parameters and forwards them to the control software for processing.

## III. SOFTWARE DEVELOPMENT

The system software employs a client-server architecture for robot control and feedback information display. A brief functional software structure of the platform is shown in fig.4.

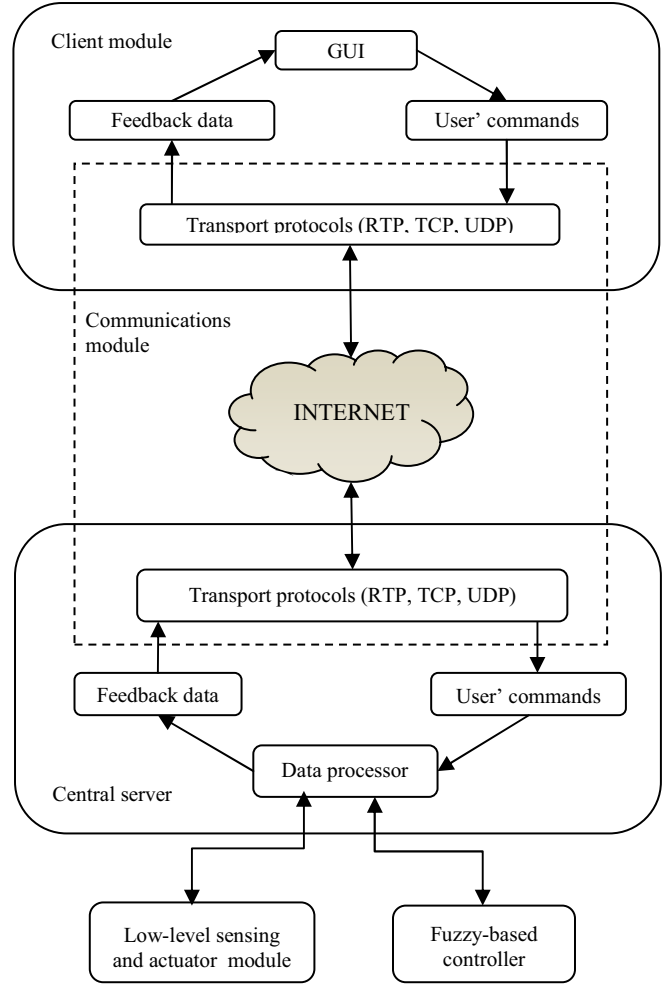

Figure 4. System software architecture

In the model, there are four modules: the communication module, the central server, the fuzzy-based controller, and the client module. Details of these modules are described as follows.

*A. Multi-protocol communications*

Various types of information need to be exchanged between the robot and the human operator. Generally, there are three classes of data:

- Administrative data (such as access control, user validation, and configuration data) and user control commands (such as desired translation velocity and rotation angle): small packet size, non-periodic transmission and requiring reliable delivery.

- Image data (the most important and costly information feedback): large packet size, periodic transmission, real-time delivery is required, requires significant bandwidth, and the most recent information is preferred should packets become lost.

- Continuous control data (joystick signal) and feedback information on the scene and the robot (such as position of the robot and sensing data): small packet size, periodic transmission, real-time delivery is required, and the most recent information is preferred should packets become lost.

From the above categorization, it is clear that all types of information require real-time delivery except for once-for-all administrative data and user control commands. Consequently, to obtain the optimal performance, different transport protocols should be used for the transmission of each information category.

There are currently three main transport protocols available for implementing remote control applications over the Internet: the User Datagram Protocol (UDP) [16], the Transmission Control Protocol (TCP) [17], and the Real-time Transport Protocol (RTP) [18].

UDP is based on the idea of sending a datagram from a device to another as fast as possible without due consideration of the network state. This protocol does not maintain a connection between the sender and the receiver, and it does not guarantee that the transmitted data packets will reach the destination as well as the chronological order of the data at the receiving end. TCP is a more sophisticated protocol which was originally designed for the reliable transmission of static data such as e-mails and files over low-bandwidth, high-error-rate networks. In each transmission session, TCP establishes a virtual connection between the sender and receiver, performs the acknowledgment of received data packets and implements the retransmission mechanism when necessary. TCP can also adapt to the variation of network condition by applying strict congestion control policy with slow start, fast recovery, fast retransmit and window-based flow control mechanisms. RTP is the standard for delivering real-time multimedia data. The protocol provides facility for jitter compensation and detection of out-of-sequence arrival in data, and it is usually used in conjunction with the RTP Control Protocol (RTCP). More details of comparison between protocols is investigated in our previous work [20].

From the above analysis, it is well recognized that RTP is suitable for video streaming; UDP has advantages in sensing data transmitting while TCP is the best in delivering administrative data and user control commands. This configuration is successfully implemented in our communication module.

### B. Central Server

In the system, the server handles all control requests from clients, processes them and forwards the translated data to the MSSR. The control requests include the user operation and autonomous movement. Users can control the remote robot by sending primitive commands and using the arrow buttons in the user interface. Users can also input commands by using a joystick attached at client computer. After receiving the commands, the robot moves towards corresponding direction until the user pushes the stop button. During the movement, the robot can exert local intelligence to avoid obstacles or to move autonomously to a pre-defined safe point if a network interruption event is detected. Dead reckoning and obstacle avoidance algorithms are involved in the local intelligence of the robot and handled by fuzzy logic-based controller.

For the feedback data, the server periodically retrieves sensor information about status of the robot and transmits it to clients. The sensor data includes the battery level, robot position and speed, ultrasonic and laser ranges, compass deflection angle and GPS longitude and latitude. The sensing data is packetized as described in fig.5 and transmitted over UDP protocol.

| Total length 16 bits | Checksum 16 bits |
|---|---|
| Sensor data separated by "::" string ||

Figure 5. Packet format of data feedback at application level

In the system, the server program was written in Visual C++ and the communication between the server and client was established through socket, an abstraction that represents a terminal for communication between processes across a network.

### C. Fuzzy Logic Control

The objective of the fuzzy logic-based controller in our system is twofold. Firstly, it adjusts parameters of the robot during the process of user operation according to network conditions; secondly, it navigates the mobile robot to avoid obstacles and to reach a pre-defined safe point in case of a network interruption. The implementation of the fuzzy logic-based controller is divided into three circumstances:

- The fuzzy algorithm to adjust robot parameters according to network conditions.
- The fuzzy algorithm to avoid obstacles.
- The fuzzy algorithm to find the safe point.

In each circumstance, the implementation of fuzzy algorithm consists of four steps: defining the problem, defining the linguistic variables and the membership functions, defining the fuzzy rules and defuzzification. The details of fuzzy algorithms implementation were described in our previous paper [19].

### D. Graphic User Interface (GUI)

The user interface is designed with the intention of making it easy for users to interact with the mobile robot. Through the GUI, users are able to observe the remote environment, access the system parameters, and control the robot in real time over the Internet. Fig.6 shows the designed GUI which is split into four areas: the system parameters, the manual control, the visual feedback display and the virtual represent.

*The system parameters* display information about the current status of network and robot. The network status includes the connecting state, the time delay and the delay jitter. The robot parameters are the battery level, the robot position and speed, the compass and GPS data, the sonar ranging and the laser scanning data. This area also handles the function of establishment and termination of network connection.

*The manual control* displays functioning buttons for users to input commands to control the mobile robot over the Internet. A joystick is available to supply users with a more intuitive and interactive control.

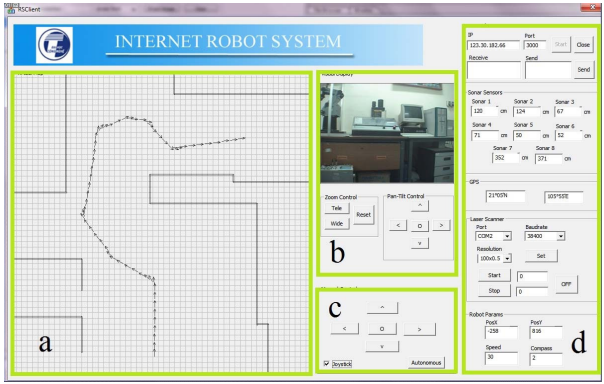

Figure 6. The graphic user interface
a) Virtual represent b) Visual feedback display
c) Manual control d) System parameters

*The Visual feedback:* The continuous and steady image stream feedback from the robot site is necessary when the Internet users control the mobile robot at the client site. Moreover, the image quality should be good enough to provide as much information about the remote site for teleoperation as possible. In this paper, H.264 algorithm [21] is used to compress the images before sending, and decompress them after receiving. H.264 is a kind of video compression scheme which supports video meeting application. It makes high grade media stream transfer through the low bandwidth network. By similar forms, the zoom and the angles of the PTZ camera can be set via the Internet. The PTZ camera can help users have more information about remote environment, especially for the mobile robot.

*The virtual represent* is a module that works on the client to handle the sensing data packets sent by the central server. Based on the extracted data, the virtual world model draws an arrow representing the robot position and direction as well as the trajectory at the specific coordinate. An environment map is built based on the sonar and laser readings, and be updated every 100ms. With this virtual environment map, the user can find suspected obstacles nearby, and make correct decision when visual feedback suffers from serious time-delay or obstacles beyond the camera's scope.

## IV. EXPERIMENTS

During the project development, different configurations were tested in different environments. The aim is to develop a more reliable system framework that can be used in the real world.

As shown in fig.7, the MSSR mobile robot was controlled from a distance of 30km to explore the laboratory we are working in; the PTZ camera was used. In another test, the MSSR was controlled from a distance of 20km to moving outdoors around the university campus while avoiding several static obstacles; the PTZ camera and the virtual represent was used in this test.

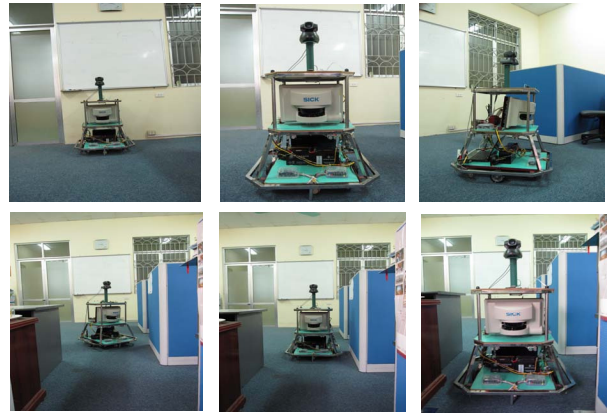

Figure 7. A sequence of images showing the motion of robot in a laboratory environment during the tele-navigation operation

Fig.8 shows the moving paths of the robot at the local site and the simulated path at the remote site in this experiment. Due to the network delay, there are slight differences between two paths but the maximum errors of 0.09m in horizon and 0.07m in vertical are acceptable for the direct control.

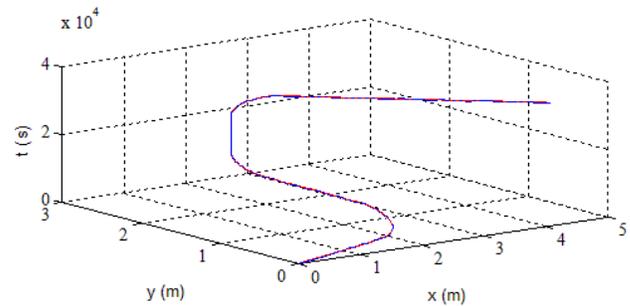

Figure 8. Moving path of the robot at the local site (blue solid line) and the simulated path at the remote site (red dashed line)

When the mobile robot moves at the low speed and few obstacles block the path, the advantage of having direct control is that the user can see the result of his/her own action without any external contribution. However, the control of the robot is more difficult in a complex environment under the high robot speed and serious network transmission delay. Table I displays time delay feature of the system network according to the transmitting data size. Due to the size of the sensing feedback data is around 500 bytes and the sampling period is 200ms, the time interval between the moment that a feedback data packet sent and the virtual represent begins to update is about 429ms. In our system, the speed of robot is from 0m/s to 1.5m/s and the time delay of the video stream is around 2000ms. A scenario in which the speed of robot is 1m/s means that users are impossible to recognize a sudden obstacle appeared in front of the robot at a distance of 0.429m by the virtual represent and at a distance 2m by the camera. The present of the autonomy is necessary in this scenario to ensure the safety of the robot and the success of navigation.

TABLE I.   TIME DELAY OF THE NETWORK

| Data Size (byte) | Time delay (ms) | | |
|---|---|---|---|
| | *Minimum* | *Average* | *Maximum* |
| 100 | 79 | 103 | 189 |
| 500 | 99 | 129 | 229 |
| 1000 | 109 | 140 | 229 |
| 2000 | 149 | 255 | 410 |

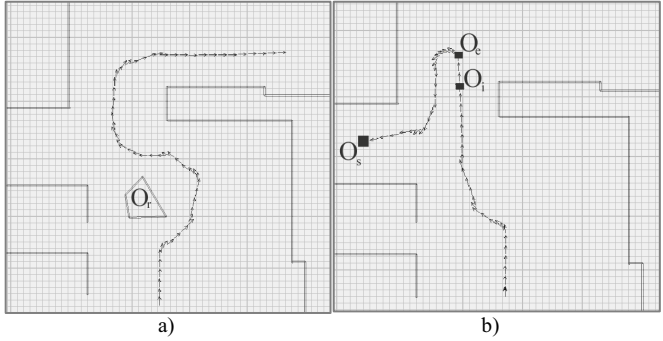

Figure 9.   Moving paths of the robot in autonomous mode
a) obstacle avoidance     b) Safe-point achievement

Fig.9a shows the experiment in which the robot himself successfully avoids the obstacle $O_r$ during a teleoperation. The situation of network interruption is investigated in a different experiment in which the operator suddenly disconnects the Internet connection at point $O_i$ of a tele-guidance process. The robot continues to move for 5s to point $O_e$ before it detects the interruption event, activates the autonomous mode and automatically navigates to the safe point $O_d$ (fig.9b).

## V. CONCLUSION

It is extremely time-consuming to build an experimental platform for the study of Internet robots from scratch. In this paper, a new modular platform for Internet mobile robotic systems is developed. The system hardware mainly consists of a Multi-Sensor Smart Robot. Many types of sensors including position speed encoders, integrated sonar ranging sensors, compass and laser finder sensors, the global positioning system (GPS) and the visual system are implemented allowing the robot to support a wide range of applications including both indoor and outdoor environment. The limitation in working area is removed by the use of 3G mobile network. The system employs a client-server software architecture for robot control and feedback information display. The exchanged data between the client and server is transmitted over the Internet by a multi-protocol model. Autonomous mechanisms based on fuzzy logic algorithms are implemented to ensure the robot safety. The platform has been tested in different environments, and the results are promising.

## ACKNOWLEDGMENT

This work was supported by Vietnam National Foundation for Science and Technology Development (NAFOSTED).